\documentclass[letterpaper, 10 pt, conference]{ieeeconf}  % Comment this line out if you need a4paper

\IEEEoverridecommandlockouts                              % This command is only needed if
\usepackage{graphics} % for pdf, bitmapped graphics files
\usepackage{epsfig} % for postscript graphics files
\usepackage{mathptmx} % assumes new font selection scheme installed
\usepackage{times} % assumes new font selection scheme installed
\usepackage{amsmath} % assumes amsmath package installed
\usepackage{amssymb}  % assumes amsmath package installed

\usepackage{amsthm}
\usepackage{mathrsfs}

\usepackage{color}
\usepackage{enumerate}
\usepackage{subfigure}
\usepackage{url}
\usepackage{tablefootnote}
\usepackage[para,online,flushleft]{threeparttable}
\usepackage[colorlinks, linkcolor=black]{hyperref}
\usepackage{multirow}
\usepackage{multicol}
\usepackage{booktabs}
\usepackage{longtable}
\usepackage{subcaption}
\usepackage{tabularx}

\title{\LARGE \bf
Integrating Controllable Motion Skills from Demonstrations
}

\author{Honghao Liao$^{1}$, Zhiheng Li$^{1}$, Ziyu Meng$^{1}$, Ran Song$^{1}$, Yibin Li$^{1}$ and Wei Zhang$^{1}$ % <-this % stops a space
\thanks{$^{1}$Honghao Liao, Zhiheng Li, Ziyu Meng, Ran Song, Yibin Li and Wei Zhang are with the School of Control Science and Engineering, Shandong University, Jinan, China.}%
\thanks{Corresponding author: Wei Zhang (Email: davidzhang@sdu.edu.cn)}
}

\begin{document}
\maketitle

% \twocolumn[{
% \renewcommand\twocolumn[1][]{#1}
% \maketitle
% \begin{center}
%     \captionsetup{type=figure}
%     \includegraphics[width=0.95\linewidth]{fig/poster.png}
%     \captionof{figure}{Our method enables legged robots to flexibly integrate a range of different motion skills into a single controller, and can be further combined with a high-level NLI module to enable preliminary language-directed skill control. The language commands used here are (1) "Act as if you're a scary character", (2) "Return to normal walking style", (3) "Show me your jumping skills".}
% \end{center}
% }]

\begin{figure*}[!htbp]
    \centering
    \includegraphics[width=0.95\linewidth]{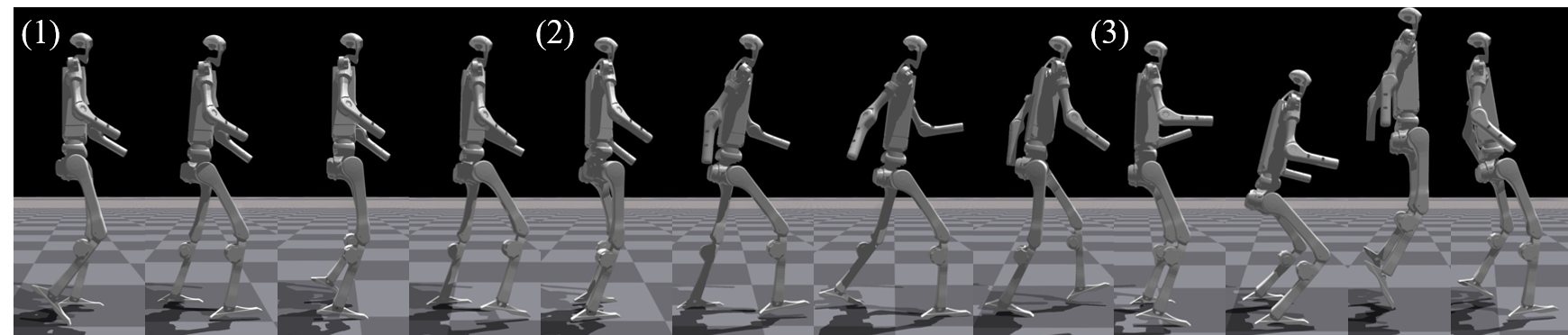}
    \caption{Our method enables legged robots to flexibly integrate a range of different motion skills into a single controller, and can be further combined with a high-level NLI module to enable preliminary language-directed skill control. The language commands used here are (1) "Act as if you're a scary character", (2) "Return to normal walking style", (3) "Show me your jumping skills".}
    \label{fig:image_0}
\end{figure*}

% \maketitle

\thispagestyle{empty}
\pagestyle{empty}

%%%%%%%%%%%%%%%%%%%%%%%%%%%%%%%%%%%%%%%%%%%%%%%%%%%%%%%%%%%%%%%%%%%%%%%%%%%%%%%%
\begin{abstract}
The expanding applications of legged robots require their mastery of versatile motion skills. Correspondingly, researchers must address the challenge of integrating multiple diverse motion skills into controllers. While existing reinforcement learning~(RL)-based approaches have achieved notable success in multi-skill integration for legged robots, these methods often require intricate reward engineering or are restricted to integrating a predefined set of motion skills constrained by specific task objectives, resulting in limited flexibility. In this work, we introduce a flexible multi-skill integration framework named Controllable Skills Integration~(CSI). CSI enables the integration of a diverse set of motion skills with varying styles into a single policy without the need for complex reward tuning. Furthermore, in a hierarchical control manner, the trained low-level policy can be coupled with a high-level Natural Language Inference~(NLI) module to enable preliminary language-directed skill control. Our experiments demonstrate that CSI can flexibly integrate a diverse array of motion skills more comprehensively and facilitate the transitions between different skills. Additionally, CSI exhibits good scalability as the number of motion skills to be integrated increases significantly.
\end{abstract}

%%%%%%%%%%%%%%%%%%%%%%%%%%%%%%%%%%%%%%%%%%%%%%%%%%%%%%%%%%%%%%%%%%%%%%%%%%%%%%%%

\section{INTRODUCTION}
\label{sec:introduction}
% In recent years, reinforcement learning-based~(RL-based) approaches have reaped considerable success in the field of motion control for legged robots, e.g., locomotion over complex terrain~\cite{takahiro2022learning}, or the impressive quadruped parkour~\cite{zhuang2023robot} \ziyu{maybe 3 instances better?}. \zhiheng{Thinking in terms of human-robot interaction, one of the questions that naturally comes to mind is whether it is possible to integrate mutiple different motion skills into a single controller and make them controllable?} However, existing RL-based approaches do not fulfil this purpose well.\ziyu{However, existing RL-based approaches do not adequately fulfil this objective well} One of the most hindering problems is the complex reward engineering required for skill learning using related methods. More frustratingly, the rewards designed for different skills are typically not generalizable, complicating the integration of multiple motion skills.

% As the use of legged robots becomes more widespread, the demand for their capabilities will continue to increase. 
With the increasing prevalence of legged robots, the demand for their enhanced capabilities is expected to grow continuously.
One significant trend is the expectation for legged robots to manage a diverse array of motion skills to potentially cope with a wide range of tasks in real-world applications. Recently, RL-based approaches have reaped considerable success in the task of multi-skill integration for legged robots, e.g., impressive quadruped parkour~\cite{zhuang2023robot}, omnidirectional bipedal locomotion~\cite{rodriguez2021deepwalk}, or wonderful bipedal robots football match~\cite{tuomas24learning}. However, these RL-based approaches do not adequately fulfill this objective. One of the most hindering problems is the complex reward engineering required for skill learning using related methods. More frustratingly, the rewards designed for different skills are typically not generalizable, complicating the integration of multiple motion skills.

% More frustratingly, the rewards that need to be designed for different skills are usually not generalisable, which makes the integration of multiple motion skills extremely difficult. \ziyu{More frustratingly, the rewards designed for different skills are typically not generalizable, complicating the integration of multiple motion skills.}

% \ziyu{Is the word "integrate" better than "combine" in our methods? }
Integrating imitation learning~(IL) with RL provides a feasible solution to these problems. Recently, a plethora of ongoing research in the field of character animation has demonstrated the effectiveness of this paradigm. By allowing agents to track the reference motion trajectories~\cite{RoboImitationPeng20,2018-TOG-deepMimic,bergamin19drecon}, or aligning with motion style from imitating reference motion~\cite{2021-TOG-AMP, wang2023learning, escontrela2022adversarial}, controllers trained via IL can perform motion skills naturally. As demonstration data serves as a reference for policy learning, RL approaches integrated with imitation learning significantly simplify reward engineering. Additionally, the design of rewards necessary for various motion skills becomes more standardized.

Recently, some methods in character animation have been adapted to the multi-skill integration for legged robots~\cite{escontrela2022adversarial, li2023learningterrain, tang2023humanmimic, vollenweider2023advanced, han2023lifelike}. However, these approaches generally exhibit limited flexibility. Most of them rely on extra well-defined task objectives~\cite{escontrela2022adversarial, li2023learningterrain, tang2023humanmimic} to integrate a set of similar task-related skills, thereby constraining their applicability and the variety of motion skills they can integrate. 
% Alternatively, other approaches~\cite{vollenweider2023advanced, han2023lifelike} attempt to integrate diverse skills by employing additional networks or training stages, which results in substantial training cost as the number of motion skills increases. 
Alternatively, other approaches~\cite{vollenweider2023advanced, han2023lifelike} attempt to integrate each different motion skill with the help of additional networks or training stages, which makes the corresponding training costs increase when the number of skills to be integrated increases. Overall, these approaches are limited in the range or number of motion skills that can be integrated.

In this work, we propose CSI, a flexible framework designed for legged robots to integrate multiple motion skills from reference motion clips into a single controller. CSI is built upon Generative Adversarial Imitation Learning~(GAIL)~\cite{ho2016generative}, an IL framework that obviates the need for skill-specific reward engineering. Furthermore, by incorporating key designs such as Conditional Imitation Learning and Condition-Aware Loss, CSI can use skill labels as a control interface for integrated motion skills, which makes it possible to access some external knowledge like natural language for skill control. Our experiments validate the effectiveness of CSI and demonstrate its notable ability to support language-directed skill control through the incorporation of high-level NLI modules. 

In summary, the primary contributions of this paper are reflected in the following three aspects:
\begin{itemize}
\item We propose CSI, a flexible multi-skill integration framework. CSI enables legged robots to acquire versatile and controllable motion skills by effectively imitating reference motion capture data.
% thus bridging the gap in \bigliao{multi-motion control} for legged robots.
% \item \textcolor{cyan}{We introduce torque control as a more efficient alternative to PD position control in our tasks. Our experiments demonstrate that torque control significantly enhances motion fluency and energy efficiency, particularly in tasks involving learning multiple motions.}
% \item By combining interaction with NLI task, we achieve open-vocabulary multi-motion interaction. This designation effectively decouples high-level neural language processing module from low-level policy, which can make full use of pre-trained NLI model and also better meet the customisation requirements in applications.

% NLP modules also can be replaced by other controller
% Proposal: enabling the open-vocabulary control

\item Our approach provides a more controllable interface, enabling the flexible and easy leveraging of heuristic knowledge to improve the efficiency of skill execution.
% \textcolor{cyan}{Todo.}
% We introduce a xxx generator 以复杂的文本作为输入，可以产生 embedding space of control 指令？ aligned with GAIL 采样的 motion embedding space, thus 可以控制 motion skills/ controllable
% \item Comprehensive experimental verification and analysis are conducted with legged robot in both simulation and reality, which underscores the effectiveness and practical applicability of our proposed methods in varied settings.
% \item Comprehensive experimental verification and analysis are conducted on three different morphological legged robot platforms, AlienGo~\cite{AlienGo}, BRUCE~\cite{BRUCE} and H1~\cite{H1}, which confirm the validity and adaptability of our proposed methods.
\item Detailed experiments and analyses of our approach are carried out on different datasets as well as on different robots, which validates the effectiveness and adaptability of our work.
\end{itemize}
Supplementary videos of this work are posted on \url{https://vsislab.github.io/CSI_IL/}.

\section{RELATED WORK}
\subsection{Multi-Skill Integration for Legged 
 Robots}
% \bigliao{Another logic (without mentioning the work in the field of CA): RL methods have made a lot of progress on multi-skill integration for legged robots -- but RL methods require reward engineering -- so some RL methods introduce IL paradigm.}

% The integration of multiple motions into one or several controllers is currently available in a number of feasible ways. By using optimization-based methods such as TO, ~\cite{Atlas} allows the bipedal robot Atlas to exhibit a wide variety of agile motions such as jump and backflip. In recent years, methods based on RL have also been more successful in integrating multiple motions. For example, \cite{rodriguez2021deepwalk} have made bipedal robots learn omnidirectional locomotion, including walk forward, walk backward and turning; another example is controllers trained by RL enable quadrupedal robots learn to creep forward, jump and other parkour motion skills~\cite{zhuang2023robot}.

% \textcolor{blue}{Early optimization-based methods have achieved the integration of several motions into one or a number of controllers.} The bipedal robot Atlas \cite{Atlas} employs the MPC and the TO methods to adapt various agile motions such as jumping and backflip. 
In recent years, methods built upon the RL paradigm have achieved promising performance in integrating multiple motion skills for legged robots. Rodriguez \textit{et al.} \cite{rodriguez2021deepwalk} introduced deep reinforcement learning~(DRL) method to enable the bipedal robot NimbRo-OP2X~\cite{ficht2020nimbro} to learn agile omnidirectional locomotion, including walk forward, walk backward and steering. Zhuang \textit{et al.} \cite{zhuang2023robot} proposed a two-stage RL training approach for the quadruped robots Go1~\cite{Go1} and A1~\cite{A1} to acquire complex dexterous parkour maneuvers, such as creep forward and jumping. Tuomas \textit{et al.} \cite{tuomas24learning} employed DRL techniques to impart agile soccer skills on the small bipedal robot OP3~\cite{OP3}.

However, RL-based methods typically require meticulous reward engineering, which can be both labor-intensive and time-consuming, especially for multi-skill integration. To address this problem, the introduction of IL has proven effective in alleviating the need for explicit reward design, thereby offering considerable advantages for integrating multiple motion skills. This beneficial impact has been demonstrated by related works in the field of character animation. Won \textit{et al.} \cite{won2020ascalable} proposed a tracking-based method to integrate various motion skills from a large-scale open-source motion capture dataset~\cite{CMU} into a few controllers, and characters can switch between controllers to perform different motion skills. Peng \textit{et al.} \cite{2022-TOG-ASE} and Zhu \textit{et al.} \cite{zhu2023ncp} incorporated a shared embedding space within the learning process, enabling the integration of multiple skills into one single policy. Furthermore, this embedding space also serves as an interface to call integrated skills for subsequent training of downstream tasks.

% \zhiheng{need a conjuction} Some of the related methods in character animation has also been migrated to the field of legged robots recently. 

These methods have also witnessed initial explorations on legged robots in recent studies. \cite{escontrela2022adversarial, li2023learningterrain, tang2023humanmimic} incorporated velocity command tracking objectives during the training process to integrate a set of similar walking and running maneuvers into one controller. Vollenweider \textit{et al.} proposed MultiAMP~\cite{vollenweider2023advanced}, which assigns an additional discriminator for each skill that needs to be integrated to aid learning. Han \textit{et al.} \cite{han2023lifelike} instead adopted a multi-stage training process, training specific Vector Quantized-Variational AutoEncoder~\cite{van2017neural} policy for each different skill, then integrating these policies into a single one by distillation.
% \bigliao{At the beginning, the first thing should be stated is that all related methods in the field of CA are only for virtual characters and a lot of privileged elements (e.g. privileged observation, residual force, etc.) are introduced when integrating multi-skills. In the field of legged robots, however, it is often impractical to introduce these privileged elements. Therefore, some recent approaches have been adapted for these aspects and transferred to legged robots applications.}

% A further question that easily comes to mind after the integration of multiple motions is how to control these integrated motion skills.
\subsection{Controllable Motion Skills}
How to give controllability to the integrated skills is an important issue for the integration of multiple motion skills in legged robots. \cite{escontrela2022adversarial, li2023learningterrain, tang2023humanmimic, li2023learning} added additional task objectives such as tracking velocity command, to the training objective as a way to achieve controllability of the integrated motion skills through command input. This approach is constrained by the need for well-defined task objectives and is also limited by the specific task requirements that determine which motion skills can be integrated. For example, when the task involves velocity tracking, it becomes more difficult to integrate dance movements into the policy. Vollenweider \textit{et al.} ~\cite{vollenweider2023advanced} applied one-hot skill code as input, each code corresponds a skill discriminator, through leanring to switch between these codes, the policy can adjust the output motion skill according to the input skill. 
Although this approach allows for a less restricted range of integrable motion skills, the training cost increases proportionally with the number of skills that need to be integrated.

% A crucial question arises regarding the control of these integrated motion skills. \zhiheng{too slight}
% Most of current approaches~\cite{2018-TOG-deepMimic, 2021-TOG-AMP, escontrela2022adversarial, 2022-TOG-ASE, vollenweider2023advanced, peng2019mcp} rely on additional task training objectives (e.g. velocity commands tracking) to incentivize the policy to generate motions that align with user command input, thereby achieving controllability over the integrated motions. While effective, this approach is limited to scenarios where the task goal is clear and well-defined, which may not be practical in certain cases. Alternatively, other methods~\cite{2022-SA-PADL, tessler2023calm, dou2023case} offer user-friendly control interfaces for integrated motions, such as natural language and state machine, leveraging shared embedding space and multi-stage training. However, the high training cost associated with these methods, attributed to the need for multi-stage training or embedding processes, restricts their practical applicability.

In this paper, we propose CSI, a flexible framework that enables the integration of diverse motion skills into a single controller. CSI uses skill labels as control signals to enable controllability through integrated motion skills, which can be further combined with a pre-trained NLI module to achieve preliminary language-directed skill control. Unlike existing approaches, CSI eliminates the need for additional task objectives and multi-stage training to establish a control interface. We anticipate that our work will serve as a valuable reference for future multi-skill integration applications in legged robots.

\begin{figure*}[t]
\centering
\includegraphics[width=0.95\linewidth]{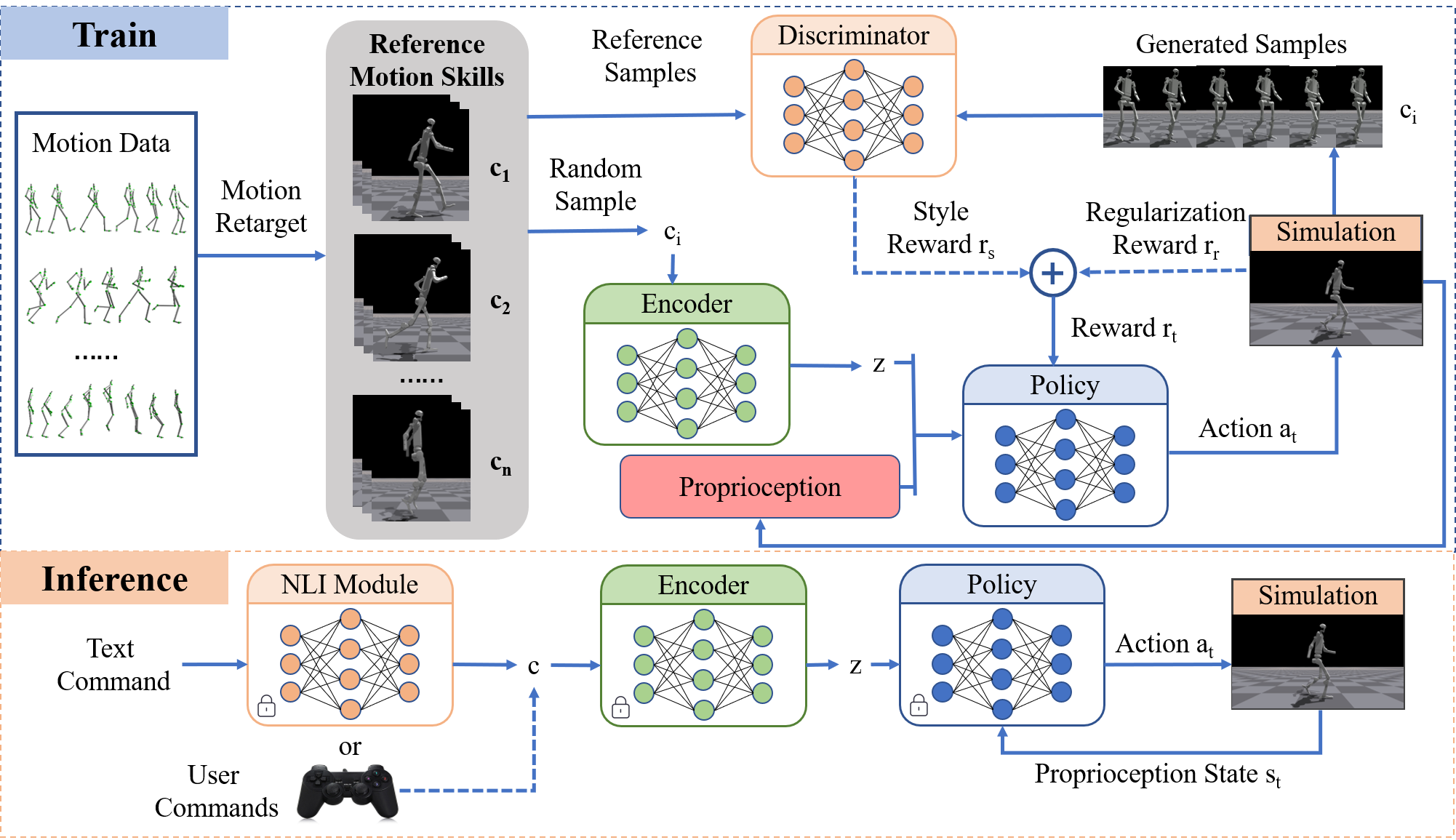}
\caption{Overview diagram of CSI. Through retargeting and skill labeling, a set of reference motion clips with corresponding labels can be obtained. During training, sampled motion skill labels $c_i$ are mapped to latent vectors $z$ through an encoder network, and the policy generates corresponding motion skills based on $z$. The discriminator is responsible for indirectly regulating the motions generated by the policy in a way that provides style rewards. After the training stage, a controller with integrated multiple motion skills is available. These integrated skills can be controlled directly through user commands or externally via a high-level pre-trained NLI module for language-directed skill control.}
\label{fig:framework}
\end{figure*}

\section{METHOD}
\subsection{Preliminary}
In this work, our goal is to enable legged robots to obtain versatile and controllable motion skills. To achieve this, we formulate the problem as a goal-conditioned~\cite{chane2021goal} Markov Decision Process~(MDP) $(S, A, $c$, R, p_0, \gamma)$, where $S$ is the state space, $A$ denotes the action space, $c \in C$ signifies the input condition, $R = r(s_{t}, s_{t+1}, c)$ represents the reward for each time step, $p_0$ is initial state distribution, and $\gamma \sim (0, 1]$ denotes the discount factor. During the training process, we employ the RL algorithm to optimize the parameters of policy: $\pi_\theta: S \to A$, aiming to maximize the expected return of the discounted episode reward $J(\pi)=E_{c \in C, \pi_\theta} [\sum_{t=0}^{T-1} \gamma^{t} r_t]$, where $T$ represents the horizon length of an episode.

To enable legged robots to generally learn motion skills by imitating from demonstration motion data, our method is built upon GAIL~\cite{ho2016generative}. In GAIL's framework, a generator is trained in an adversarial manner with a discriminator. The discriminator's role is to distinguish between \emph{real} samples from expert demonstrations and \emph{fake} samples generated by the generator. The feedback from the discriminator to the generator serves as a reward signal, guiding the generator to produce data that closely resembles the expert demonstrations. Typically, GAIL requires state-action pairs $(s_{t}, a_{t})$ as input, where expert demonstration data provides both the state and the corresponding action. In our approach, we utilize motion capture data as the source of expert demonstration. Motion capture data primarily records the skeletal states at each frame but does not include the explicit actions taken by the expert. To address this limitation, we adopt the paradigm of GAILfO~\cite{torabi2019adversarial} by utilizing state transitions $(s_{t}, s_{t+1})$ instead of state-action pairs $(s_{t}, a_{t})$. This adaptation enables the generator to learn from the evolution of states, even in the absence of explicit action conducted by the expert.

\subsection{Conditional Imitation Learning} \label{CLI}

To achieve controllability over the integrated motion skills, we propose that all networks within the framework should be guided to operate according to some kind of instruction. Therefore, we introduce Conditional Imitation Learning~(CIL) to guide the networks to the specified motion skills during the training process. For the discriminator, we add skill labels to the original samples input, which requires it to be able to judge the authenticity of given samples by taking into account the skill label information. For the policy and the value function, we introduce a simple encoder network to map the skill label to a latent vector $z$. $z$ will be used as part of the input to the policy and the value function, which requires them to be able to respond according to the condition $z$. Based on the background of CIL, we design the basic training objective of the discriminator as follows:
% Moreover, it's noteworthy that the implementation of conditional imitation learning obviates the need to design separate evaluation networks for each motion skill, thereby enhancing the \bigliao{scalability} of our approach compared to the multi-discriminator style~\cite{vollenweider2023advanced}. 

\textbf{Conditional Imitation Loss.} In the framework of the vanilla GAN~\cite{goodfellow2014generative}, a variational approximation of the Jensen-Shannon divergence~\cite{menendez1997jensen} is commonly applied to achieve the adversarial training objective. Building on the concept of conditional probability, we introduce conditions based on the original objective:
\begin{equation}
\begin{split}
    L_{I} = &-E_{(s_{t}, s_{t+1}) \in d^{M}}[log(D(s_{t}, s_{t+1} | c))] \\
    &- E_{(s_{t}, s_{t+1}) \in d^{\pi}} [log(1-D(s_{t}, s_{t+1} | c))] \\
\end{split}
\end{equation}
where $d^{M}$ and $d^{\pi}$ represent the state transition distributions of the reference motion skills and those generated by the policy, respectively. $c$ denotes the corresponding skill label. Given a motion state transition $(s_{t}, s_{t+1})$ combined with the corresponding skill label $c$, the output of the discriminator is expressed as $D(s_{t}, s_{t+1} | c)$. 
% This improvement is equivalent to setting separate training objectives for different motion skills, which significantly mitigates mode collapse.

\textbf{Gradient Penalty.} We also introduce gradient penalty, which has been proven effective in reducing the destabilizing effects of adversarial training~\cite{2021-TOG-AMP,gulrajani2017improved7,mescheder2018training}:
\begin{equation}
    L_{GP} = E_{(s_{t}, s_{t+1}) \in d^{M}}[\left\| \triangledown D(s_{t}, s_{t+1}) \right\|^{2}_{2}]
\end{equation}
Note that we calculate the gradient penalty with respect to all real samples, irrespective of the conditions corresponding to those samples.

In this way, we can preliminarily guide the generator to mimic the specified motion skills during the training process. However, some practical considerations deserve to be noticed. For example, since training directly based on the above training objectives is still an unsupervised paradigm, input conditions are frequently disregarded by the networks during training, which results in controllers that exhibit only a limited set of uncontrollable motion skills. We add some additional designs to the training objectives of the discriminator to cope with these problems:

% However, merely adding conditions to the input is insufficient. We observe that input conditions are frequently disregarded by the networks, resulting in a controller that exhibits only a limited and monotonous set of uncontrollable motion skills. Therefore, we propose enhancements to the training objective: 

\textbf{Condition Aware Loss.} Due to the unsupervised learning nature of GAN, despite the conditional imitation loss described above, the discriminator still tends to ignore the input conditions during the training process, especially when the number of reference motion skills increases. To enhance the discriminator's sensitivity to motion skill labels, we construct mismatched samples that carry mismatched motion skill labels, alongside real and fake samples. In our setting, the mismatched samples should be judged as negative by the discriminator:
\begin{equation}
    L_{CA} = -E_{(s_{t}, s_{t+1}) \in d^{M}} [log(1-D(s_{t}, s_{t+1} | \hat{c}))]
\end{equation}
where $\hat{c}$ denotes the skill labels that do not match the input samples. Subsequent experiments demonstrate that $L_{CA}$ significantly mitigates mode collapse and enables the controller to comprehensively master the integrated motion skills.

\textbf{Weight Decay.} For multi-skill integration, the reference sample capacity is typically limited, rendering the GAN network prone to overfitting and leading to a relative lack of diversity in the skills generated by the final trained controllers. To alleviate this issue, weight decay is introduced for the discriminator:
% The motion capture data itself is usually not perfect, some detail flaws in it, such as motion jerks and foot overhangs, may be captured by the discriminator and overly relied upon to differentiate the generated data from the reference data. This overfitting phenomenon forces the policy to learn these defects, which affects the effectiveness of the final generated motion. To mitigate this phenomenon, we introduce weight decay loss:
\begin{equation}
    L_{WD} = \Sigma || \omega_{D} || ^ {2}
\end{equation}
where $\omega_{D}$ denotes the weight parameters of the discriminator network. This improvement enables the discriminator to focus more on the general features of each skill, thereby increasing the diversity of the generated skills.

Finally, our training objective for the discriminator is defined as:
\begin{equation}
    L_{D} = \omega_{i} L_{I} + \omega_{ca} L_{CA} + \omega_{wd} L_{WD} + \omega_{gp} L_{GP}
\end{equation}
where $\omega_{i}$, $\omega_{ca}$, $\omega_{wd}$ and $\omega_{gp}$ are hyperparameters to balance each item of the training objective.

\subsection{Reward Setting}
Similar to GAIL, the reward feedback in CSI is also derived from the discriminator. To encourage the generated motion skills to resemble the reference motion capture data, while maintaining alignment between the generated motions and the given conditions, we define conditional style reward~\cite{2021-TOG-AMP} as follows:
\begin{equation}
    r_{s}=-log[1-D(s_{t}, s_{t+1} | c)]
\end{equation}
% since adversarial imitation learning can obtain motion priors from the reference motion capture dataset, this reward can not only regulate the behaviour of the policy using these priors, but also encourage the policy to generate corresponding motion skills based on given conditions.

% However, our training dataset only contains motion capture clips corresponding to multiple different motions, which reference motions transferred between various different motion skills are not included. Additionally, the style reward only introduces motion prior for each motion skill itself, which means necessary supervision for transformations between different motions is absent. We have found that this can easily lead to unnatural behaviors of the policy, such as high-frequency motion jerks, which is very detrimental for real-world deployment.

Conditional style reward provides great regularization for learning motion skills under the distribution of reference motion skill dataset. However, transitions between different motion skills that are not represented in the reference dataset often result in unnatural phenomena, such as jittering. To alleviate this, we introduce some additional regularization terms:
\begin{equation}
\begin{split}
    &r_{v} = \sum \left\| \dot{q}_{t} - \dot{q}_{t+1} \right\|^{2}_{2} \\
    &r_{ep} = \sum \left | \tau_{t} \dot{q}_{t} \right | \\
    &r_{a} = \left\| a_{t} - a_{t-1} \right\|^{2}_{2}\\
    &r_{t} = \sum \left | \tau_{t} \right |
\end{split}
\end{equation}
where $q$ and $\tau$ represent joint rotation angles and torque, respectively. $\dot{q}_{t}$ denotes the angular velocity of each joint at time step $t$, $a$ means action output by policy. All symbols $\sum$ denote the summation over every Degree of Freedom~(DoF) of the robot. These regularization terms lead to a smoother overall performance of the motion skills generated by the policy.

Finally, the total reward is computed as the weighted sum of the style reward and all regularization terms: 
\begin{equation}
    r_{r} = w_{s} r_{s} + w_{v} r_{v} + w_{ep} r_{ep} + w_{a} r_{a} + w_{t} r_{t}
\end{equation}
where $w_{s}$, $w_{v}$, $w_{ep}$, $w_{a}$, and $w_{t}$ are weights used to balance each component of the rewards.

\subsection{Language-Directed Skill Control}\label{combination}
% Once the controller has integrated multiple motion skills, how to achieve controllability over these skills becomes the next focus point. To achieve open text interaction, we combine the natural language processing module with the policy in hierarchical way, where the high-level natural language processing module maps open text input into motion control commands, while the low-level policy is responsible for mapping motion control commands into target motor positions.

% Once the controller has integrated multiple motion skills, our approach can be easily combined with NLP modules for downstream open vocabulary control or human-robot interaction tasks by the way of hierarchical control.

% \textbf{Open Vocabulary Control.} The low-level controller trained by CSI provides skill control interface for high-level NLP module. In order to implement open vocabulary motion control, we choose to model the control problem of skills as a zero-shot textual entailment inference problem:

The low-level controller trained using CSI provides a skill control interface to leverage external knowledge. To implement language-directed skill control, we first manually bind an additional caption to each skill label as \textit{skill caption label}, e.g. "Walk Forward", "Sprint", "Jump", etc. Then we align the output of the NLI module with those skill caption labels in a zero-shot manner:

% a zero-shot textual entailment inference task:

Given a text input ${T_{i}}$ as the premise, and a finite set of skill caption labels ${L = \{L_{1}, L_{2}, L_{3}, ..., L_{n}\}}$ as a set of hypotheses, where ${n}$ corresponds to the number of motion skills integrated by the low-level controller. The high-level NLI module is acquired to determine the textual entailment relationship between the premise ${T_{i}}$ and each skill caption label hypothesis ${L_{i} \in L}$: entailment~(positive), contradiction~(negative) or neutral. 

% The complete processing flow is shown in Figure \textcolor{cyan}{todo}. 

After the above process, the NLI module will output the entailment scores corresponding to each skill caption label, where the skill label corresponding to the highest-scoring skill caption label will be taken as the condition input for the low-level controller. As demonstrated in the \makebox{Section~\ref{experiment}}, this zero-shot classification paradigm decouples the high-level and the low-level modules, enabling a flexible combination between these two modules.

\subsection{Implementation Details}

\textbf{Model representation.} 
The policy $\pi$, the value function $V(s_{t},c)$ and the discriminator $D(s_{t},s_{t+1},c)$ are all parameterized as shallow MLP networks, each with hidden layers of size $[512, 256]$ and rectified linear unit~(ReLU) activation functions. The encoder network is a MLP network of size $[128, 128]$, with the size of latent vector $z$ being set to fixed 8-dimensional.

\textbf{Observation space.} 
An appropriate observation representation can guide policy training effectively. In our framework, the observation of the discriminator can be represented as $\{s_{t}, s_{t+1}, c\}$, where $s_{t}$ denotes the motion state of the robot at time $t$ and $c$ denotes motion skill label. Specifically, motion state $s$ is defined in terms of a relatively complete state representation:
\begin{equation}
    s = \{h, R_{r}, v_{r}, \omega_{r}, q_{j}, \dot{q}_{j}, p_{i}\}
\end{equation}
where $h$ denotes the height of the root relative to the ground, and the root is roughly located near the center of the pelvis for humanoid robots and the geometric center of the torso for quadruped robots; $R_{r}$ represents the orientation of the root; $v_{r}$ and $\omega_{r}$ mean the linear velocity and angular velocity of the root, respectively; $q_{j}$ and $\dot{q}_{j}$ are joint position and joint velocity of the $j-th$ joint respectively; $p_{i}$ means the position of the $i-th$ end-effector expressed in the local coordinate frame of the root. 

For the policy and the value function, the observation can be represented as $\{a_{t-1}, s_{propri}, z\}$, where $a_{t-1}$ means the action of the last time step, $z$ is the vector obtained by mapping the skill condition $c$ by the encoder network, $s_{propri}$ denotes proprioceptive states, which is defined as:
\begin{equation}
    s_{propri} = \{v_{r}, g_{pro}, q_{j}, \dot{q}_{j}\}
\end{equation}
where $g_{pro}$ is the projected gravity vector, which contains information about the robot's orientation.

% Table~\ref{table:observation} details the observations and dimensions employed by the discriminator, policy and value function, illustrated using H1 as an example. For the discriminator's observations, similar to~\cite{2021-TOG-AMP}, we aggregate the three-DoF joints like hip and shoulder into spherical joints, and employ tangent-normal encoding for their rotational quantities. For single-DoF joints, rotation angles of joints are used directly.

% to ensure consistency with the motion capture data, we aggregate the three DoFs of hip joints into spherical joints. Furthermore, we employ tangent-normal encoding for all rotational quantities, except for joints with only one DoF, due to the excellent continuity and uniqueness properties of this rotation representation~\cite{2021-TOG-AMP}. For the observations of policy and critic, we directly use rotation angles for all joints instead.

% we instead used quaternions for the rotation of the root, and for the joint rotations we chose the joint rotation angles directly.

\textbf{Action space.} 
The action space of the policy is defined by the target joint rotation angles. A PD position controller translates the output of the policy into the motor torques, following the equation $\tau = k_{p} (a_{t} - \theta_{t}) - k_{d} \dot{\theta_{t}}$. In our settings, the policy is queried at a frequency of $50$ Hz, and the PD position controller operates at a frequency of $200$ Hz.

\textbf{Motion retarget.} 
For humanoid robots, we employ a retargeting method similar to the one provided in IsaacGymEnvs~\cite{makoviychuk2021isaac}, and for quadruped robots, we utilize the processing flow described in~\cite{RoboImitationPeng20}. The retargeted motion capture data are collected as reference dataset $D^{M}$ for the subsequent training. State transitions sampled from $D^{M}$ are treated as real samples for training the discriminator. 

% stop here
\textbf{Training details.} 
To speed up the training process, we employ a distributed implementation of PPO~\cite{schulman2017proximal} across $4096$ parallel simulated environments in Isaac Gym~\cite{makoviychuk2021isaac, rudin2022learning}. The networks are trained for $2$ billion environment steps, equivalent to approximately $2$ years of simulation data, which can be collected in about $15$ hours on a single RTX TITAN GPU. See TABLE II in the supplementary for detailed training settings.

% For each training iteration, we collect a batch of around 67,000 state transitions $(s_{t}, s_{t+1})$, the value function is updated with target values computed using $TD(\lambda)$~\cite{sutton1998introduction}, and the policy is updated using advantages computed using $GAE(\lambda)~$\cite{schulman2015high}. We automatically tune the learning rate to maintain a desired KL divergence of $KL^{thresh}=0.015$ using adaptive LR scheme proposed by~\cite{schulman2017proximal}.

\section{EXPERIMENT} \label{experiment}
In this section, we conduct detailed experiments on our CSI method. To demonstrate the adaptability of CSI on different robots, we select three different legged robots for our experiments including quadruped robot AlienGo~\cite{AlienGo}, small humanoid robot BRUCE~\cite{BRUCE} and full-size humanoid robot Unitree H1~\cite{H1}. Factors such as morphology, mass, and varying numbers of DoF differ across these three robots, posing significant challenges for the controllable multi-skill integration. \makebox{TABLE~\ref{table:robot_properties}} shows some of the critical parameters of each robot. For humanoid robots, we define three different tasks: versatile locomotion~\textit{(H-Locomotion}), walking in different styles~(\textit{H-WalkStyle}), and simple interactions~(\textit{H-Interaction}). For quadruped robots, since there is little motion capture data available, we define only one locomotion task containing different gaits~(\textit{Q-Locomotion}). All of the motion capture data we use are retargeted from the CMU Mocap dataset~\cite{CMU} or specialized dataset provided by~\cite{zhang2018mode}. \makebox{TABLE I} in the supplementary shows detailed statistics on the content of the provided motion capture dataset.

\begin{table}[t]
\caption{Properties of different robots.}
\begin{center}
\begin{tabular}{|c|c|c|c|}
\hline
\label{table:robot_properties}
\textbf{Property} & \textbf{\begin{tabular}[c]{@{}c@{}}BRUCE\\ (w/o arms)\end{tabular}} & \textbf{AlienGo} & \textbf{\begin{tabular}[c]{@{}c@{}}H1\\ (w/o hands)\end{tabular}} \\ \hline
Degrees of Freedom & 10 & 12 & 19 \\ \hline
Number of Links & 11 & 18 & 22 \\ \hline
Stand Heights~(m) & 0.7 & 0.6 & 1.8 \\ \hline
Total Mass~(kg) & 4.8 & 21.43 & 48.58 \\ \hline
\end{tabular}
\end{center}
\end{table}

The following four methods are used to comparatively validate our method:
\begin{itemize}
    % \item AMP: This method allows agents to imitate learned reference motion skills and has been fully applied and validated for both simulation and real-world settings on quadrupedal robots~\cite{wang2023learning, escontrela2022adversarial, wu2023learning, vollenweider2023advanced}.
    % \item Conditional Adversarial Motion Prior~(CAMP): AMP~\cite{2021-TOG-AMP} trained with \textit{one-hot} skill labels.
    \item Conditional Adversarial Motion Prior~(CAMP): Based on AMP~\cite{2021-TOG-AMP, escontrela2022adversarial}, one-hot skill labels are used as conditional inputs to the policy, the value function and discriminator, which enables the integration of several different motion skills into a single controller.
    \item Conditional Adversarial Latent Models~(CALM)~\cite{tessler2023calm}: \textit{state-of-the-art} multi-skill integration method in the field of character animation, which achieves integration and controllability of multiple motion skills by introducing additional motion encoder and latent space.
    \item Baseline-I: Our method without Condition Aware Loss.
    \item CSI~(ours): The methodology presented in this paper.
\end{itemize}
See TABLE II and TABLE III in the supplementary for detailed settings for CAMP and CALM. Our comparative analysis primarily focuses on three key aspects: \textbf{skill coverage}, \textbf{skill controllability}, and \textbf{language-guided skill control}. Overall, through our experiments, we aim to answer the following questions:
\begin{itemize}
    \item Whether the controllers trained using CSI are capable of generating versatile and controllable motion skills?
    \item How do the policies trained with our method perform in simulation quantitatively and qualitatively?
\end{itemize}

% Due to the significant training costs when the number of integrated motions increase, we did not use MultiAMP~\cite{vollenweider2023advanced} as one of the baselines. In simulation, our comparisons focus mainly on three aspects: \textbf{skill coverage},\textbf{skill controllability} and \textbf{open vocabulary control or interaction}. Since the policies trained directly by AMP do not provide interaction interface for integrated skills, the baseline AMP do not involved in experiment of open text interaction. Overall, we hope that the experiments will answer the following questions:
% \begin{itemize}
%     \item Whether policies trained using CSI are able to generate versatile, natural and controllable motion skills?
%     \item How do the policies trained with our method perform quantitatively and qualitatively in both simulated environment and real-world deployments?
%     % \item In what ways does torque control outperform PD position control in multi-motion integration tasks?
% \end{itemize}

% \subsection{Performance in Simulation}
\subsection{Versatile Skills Integration}
CSI can be applied to various legged robots to integrate different controllable motion skills. Fig. 2 in the supplementary qualitatively illustrates some of the motion skills integrated by the controllers for different tasks. It can be seen that CSI does not necessitate a high degree of stylistic similarity among the integrated motion skills. Both similar skills (e.g., pace and trot) and more distinct skills (e.g., dance and wave hello) can be seamlessly integrated by CSI. Furthermore, due to the incorporation of the IL paradigm, CSI does not require specific reward engineering for each motion skill. These characteristics enhance the generality of CSI for multi-skill integration tasks in legged robots.

\begin{figure}[t]
\centering
\includegraphics[width=0.9\linewidth]{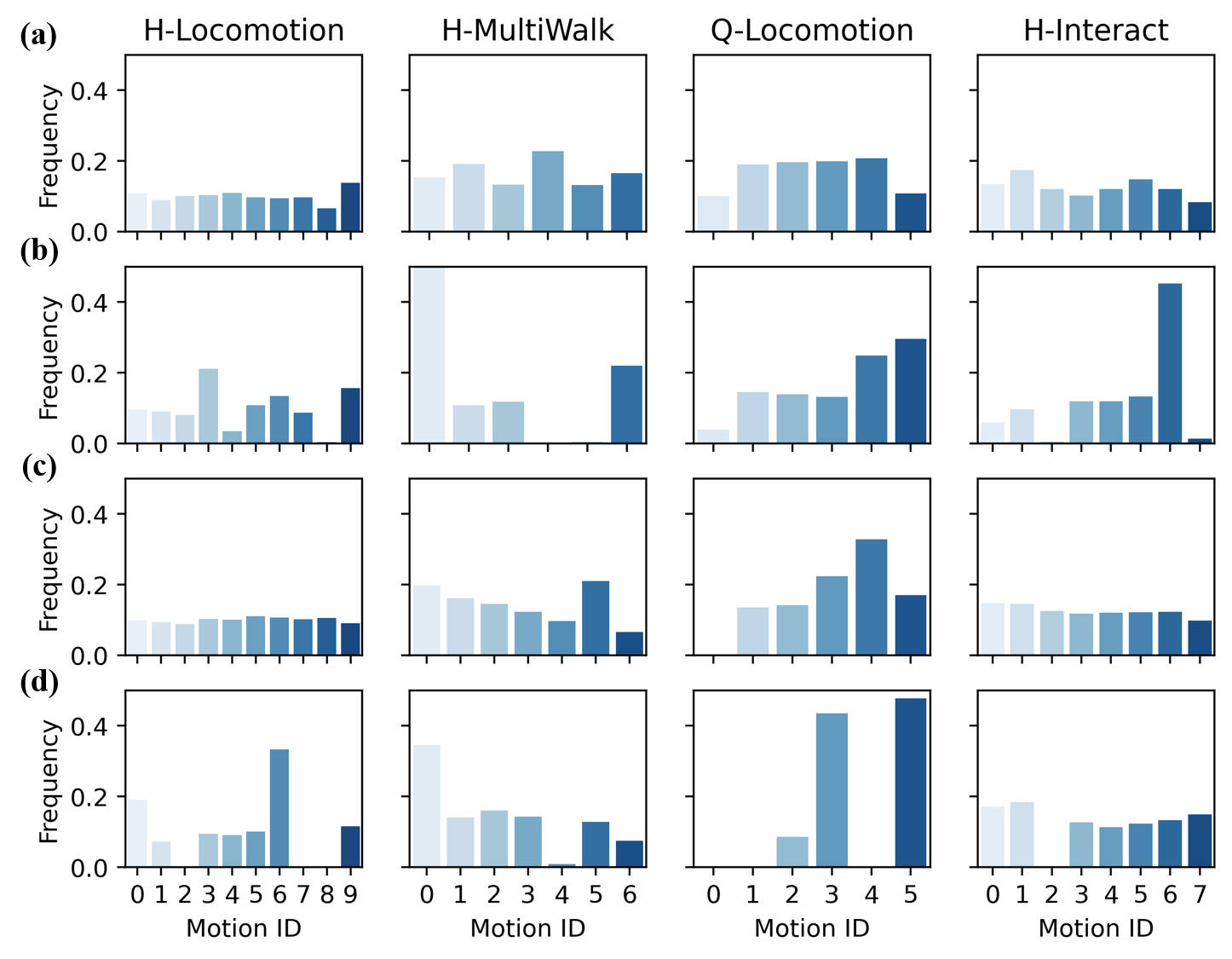}
\caption{Frequency distribution obtained by matching the controller-generated motion skills with the reference dataset under different tasks. Compared to the baselines \textbf{(b)} Baseline-I, \textbf{(c)} CAMP and \textbf{(d)} CALM, \textbf{(a)} CSI exhibits a \textbf{more even} distribution of motion skill coverage across different tasks.}
\label{fig:coverage}
\end{figure}

% \textbf{Skill coverage.} 
The following experiments will quantitatively evaluate the performance of CSI's multi-skill integration. As an integrated skills library, it should encompass all reference motion skills. Therefore, we first validate the policy trained using CSI in terms of motion skill coverage. Specifically, a motion trajectory $\tau$ is obtained by conditioning the policy $\pi$ with a randomly selected skill label $c \in C$. For each state transition pair $(s_{t}, s_{t+1})$ in trajectory $\tau$, we apply motion matching to identify the motion clip $m^{*}$ that contains the most similar motion transition pairs from the reference motion capture dataset $D^{M}$:
\begin{equation}
    m^{*} = \underset{m \in D^{M}}{arg~ min} \underset{(\bar{s}_{t}, \bar{s}_{t+1}) \in m}{min} \left \| s_{t} - \bar{s}_{t} \right \|^{2} + \left \| s_{t+1} - \bar{s}_{t+1} \right \|^{2}\label{eq:motion_matching}
\end{equation}
The process is repeated for each motion state transition pair in the motion trajectory $\tau$. The motion category that receives the highest number of matches is considered as the category of the motion trajectory $\tau$. In this experiment, a total of $2000$ motion trajectories are collected and used to validate the motion skill coverage of different methods, with each skill label sampled with equal probability. 

% Fig~\ref{fig:coverage} illustrates that CSI, compared to its baselines, is better at mitigating mode collapse that is common in GAN training paradigms, and therefore obtaining a more uniform coverage of motion skills. 

\makebox{Fig.~\ref{fig:coverage}} shows the skill coverage of each method across different tasks. Compared to Baseline-I, CSI demonstrates more comprehensive and balanced mastery of all skills in each task, highlighting the effectiveness of the supervised learning paradigm for multi-skill integration tasks. CAMP also achieves extensive skill coverage across all tasks, attributed to the advantage of the Least Square GAN~\cite{mao2017least} in mitigating mode collapse compared to the vanilla GAN. Additionally, CALM exhibits significant degrees of mode collapse across all tasks, failing to integrate some skills into the controller. We attribute this primarily to its unsupervised learning paradigm.

In addition, the flexibility to switch between different motion skills is a crucial aspect of multi-skill integration. Specifically, a comprehensive and well-balanced probability distribution of skill transition is of significant interest for controllability. To evaluate this, we allow CSI and compared methods to generate motion trajectories by randomly sampling skill label $c_{1}$ at the first $200$ time steps, followed by another skill label $c_{2}$ for the subsequent $200$ time steps. For CALM, we instead switch between motion skills by randomly sampling reference motion clips and feeding them into CALM's motion encoder to obtain new latent codes.

\begin{figure}[t]
\centering
\includegraphics[width=0.95\linewidth]{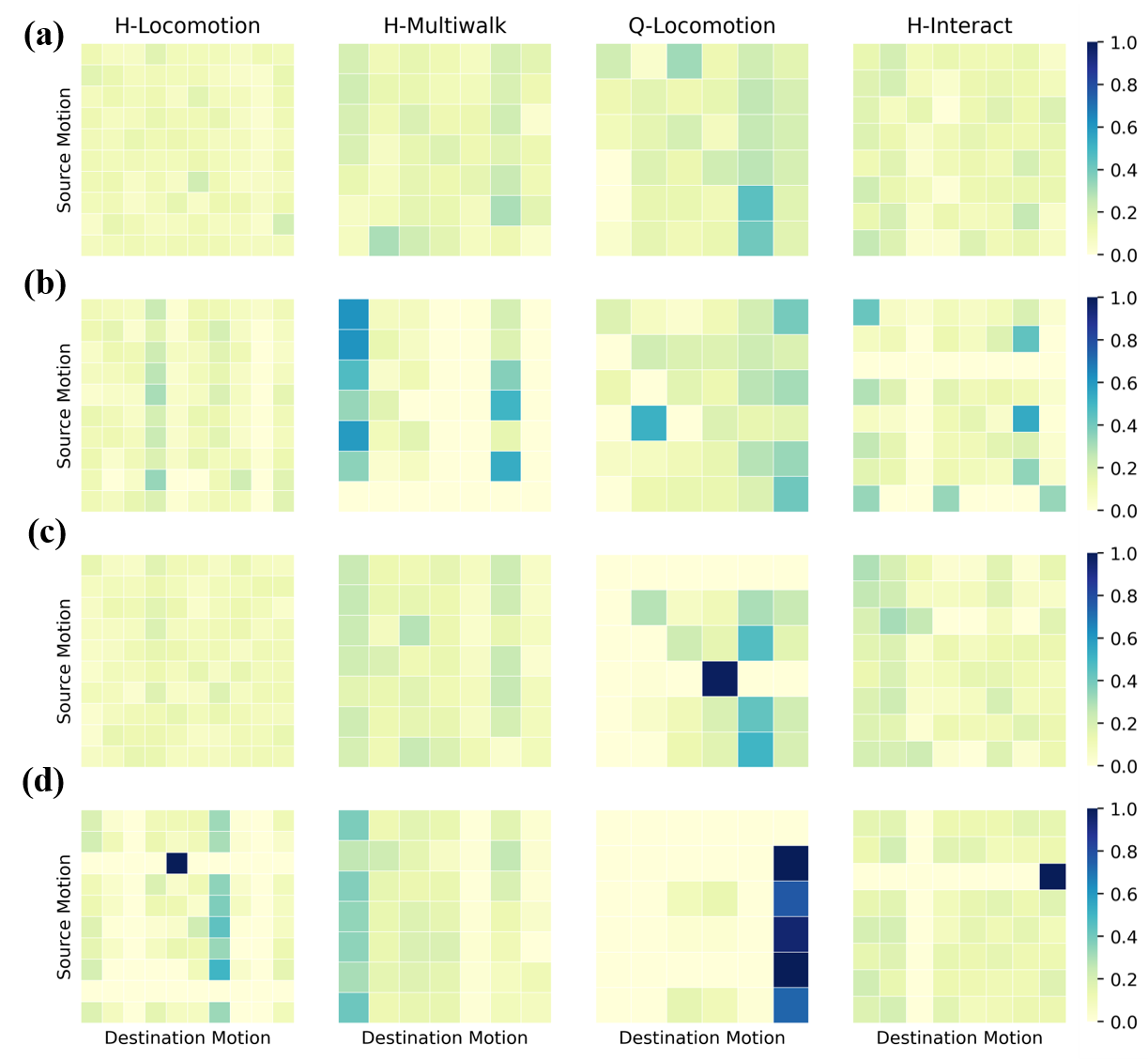}
\caption{Probability distributions of different motion skill transitions, where each row represents the probability of transferring from a source motion to each destination motion. Compared to the baselines \textbf{(b)} Baseline-I, \textbf{(c)} CAMP and \textbf{(d)} CALM, \textbf{(a)} CSI captures a \textbf{more balanced} distribution of motion skill transitions.}
\label{fig:prob_dist}
\end{figure}

Following this practice, a total of $2000$ trajectories are collected. \makebox{Eq.~\ref{eq:motion_matching}} is employed to determine the motion categories of the two motion trajectories that belonged before and after switching. \makebox{Fig.~\ref{fig:prob_dist}} illustrates the skill transition probability distribution of the four methods on each task. It can be seen that CSI exhibits a more balanced capability to switch between different motion skills compared with its baselines. 
% We further quantified the extent of coverage of skill transitions using formula $coverage~rate = \frac{transitions~from~model}{all~possible~transitions}$~\cite{2022-TOG-ASE}. Table~\ref{table:coverage} indicates that CSI has better skill transition coverage rate across different tasks compared to its baselines.

\begin{figure*}[htb]
\centering
\includegraphics[width=0.95\linewidth]{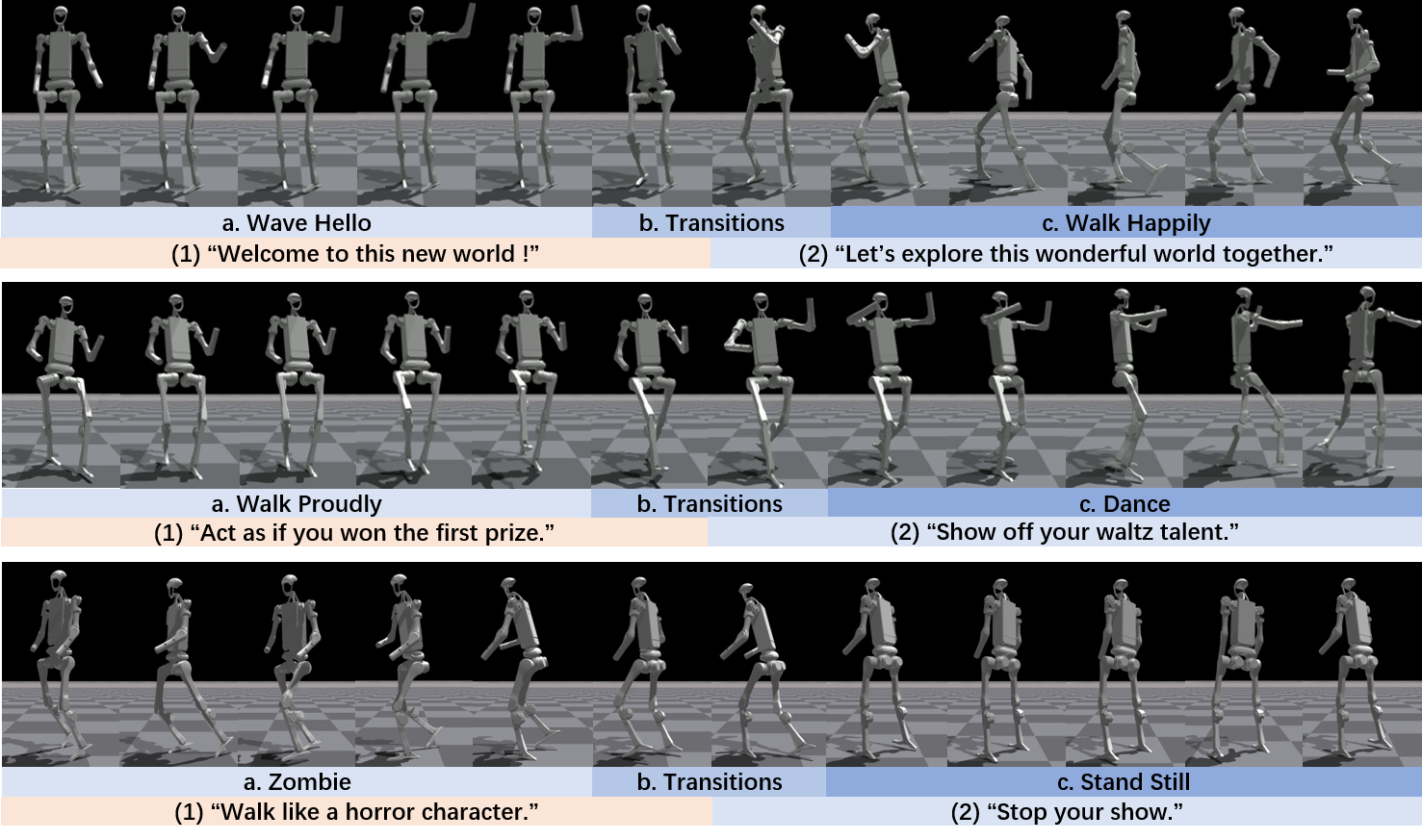}
\caption{Snapshots of the motion sequences generated by CSI, where the sub-skills (first row) and corresponding text commands (second row) are also provided. Our CSI enables language-directed skill control and transition by combining with a high-level NLI module.}
\label{fig:skill_transitions}
\end{figure*}

\subsection{Language-Directed Skill Control}
In this experiment, we qualitatively demonstrate language-directed skill control task that CSI can accomplish when combined with the pre-trained NLI module. Based on the hierarchical combination approach described in \makebox{Section~\ref{combination}}, we select the controller trained in the \textit{H-Interaction} task as the low-level module. Text descriptions from the reference motion dataset used in this task are directly employed as skill caption labels. Further details can be found in \makebox{TABLE I} in the supplementary. For the high-level module, we adapt \textit{bart-large-mnli}, a BART~\cite{mike2019bart} model trained on MultiNLI dataset~\cite{williams2018abroad}, specifically for NLI tasks. 

In each experiment, we first issue textual commands $(1)$ to \textit{bart-large-mnli}, and then switch to textual commands $(2)$ after $200$ time steps. \makebox{Fig.~\ref{fig:skill_transitions}} presents three qualitative results on the language-directed skill control task. It can be seen that \textit{bart-large-mnli} guides the low-level controller to generate semantically compliant skills based on the given textual commands, thereby facilitating the flexible switching between different motion skills.

% For the experiment of human-robot interaction, we select GPT-4 as the high-level NLP module. In each experiment, we first provide GPT-4 with selectable motion skills' range and corresponding interactive text inputs, and GPT-4 makes a comprehensive inference through the provided information to infer the next motion that most closely matches the interaction with the given text, so as to achieve simple human-robot interaction.

\subsection{Ablation Study}
% In the previous experiments, we show that one of the key designs of this work, the CA loss, has a significant impact on the performance of trained controllers. Controllers trained with CA loss have improved greatly in terms of skill coverage and motion controllability. In this section, we perform ablation study on two other key designs, the WD loss as well as DSW, to show their impact both on the training process as well as the performance of controller.

% \textbf{Condition Aware Loss.} The effect of Condition Aware Loss has been demonstrated in the previous experiments. In Figure~\ref{fig:coverage}, the baseline with no CA loss suffers severe mode collapse. Particularly in tasks \textit{H-MultiWalk} and \textit{H-interact}, the controllers tend to predominantly master only a subset of motion skills, which is clearly something we do not want to see in multi-motion integration task.

\textbf{Weight Decay.} As illustrated in \makebox{Section~\ref{CLI}}, the introduction of weight decay can increase the diversity of the generated motion skills. To quantitatively evaluate the enhancement brought by weight decay, we adopt the Average Pairwise Distance~(APD)~\cite{hassan2021stochastic,dou2023case}, which measures the diversity within a set of generated motion sequences. Specifically, given a set $M$ containing $N$ generated motion trajectories, each comprising a fixed length of $L$ frames, the APD score of $M$ is defined as:
\begin{equation}
APD(M)=\frac{1}{N(N-1)}\sum_{i=1}^{N}\sum_{j\ne i}^{N}[\sum_{t=1}^{L} (\left \| s_{t}^{i} - s_{t}^{j} \right \|^{2})]^{1/2}
\end{equation}
where ${s_{t}^{i}}$ represents the $i-th$ state of trajectory $m_{i}$. A higher APD score indicates greater diversity among the motion states of the generated motion trajectories in $M$. For each task, we set the sampling probability of different skill labels to be equal and collect a total of $2000$ motion trajectories, each with a fixed length of $200$ steps. For each experiment, we perform $10$ different samples and calculate the average APD as shown in \makebox{TABLE~\ref{table:diversity}}. Our method without weight decay is called Baseline-II. Evidently, the introduction of weight decay improves the diversity of generated motion skills on each task, which proves that weight decay effectively mitigates the overfitting of the discriminator.

\begin{table}[t]
\caption{Skill diversity of different methods, where the scores for the best performance are bolded.}
\label{table:diversity}
\begin{center}
\begin{tabular}{|c|c|c|c|c|}
\hline
Task & \begin{tabular}[c]{@{}c@{}}H-\\ Locomotion\end{tabular} & \begin{tabular}[c]{@{}c@{}}H-\\ MultiWalk\end{tabular} & \begin{tabular}[c]{@{}c@{}}Q-\\ Locomotion\end{tabular} & \begin{tabular}[c]{@{}c@{}}H-\\ Interaction\end{tabular} \\ \hline
CAMP & 1542.61 & 2023.37 & 1716.21 & 1629.36 \\ \hline
CALM & 1392.18 & 2334.03 & 1486.08 & 1790.76 \\ \hline
Baseline-I & 1634.51 & 1951.50 & 1397.09 & 1794.07 \\ \hline
% Baseline \\ (w/o CA) & 1634.51 & 1951.50 & 1397.09 & 1794.07 \\ \hline
Baseline-II & 1582.40 & 2249.88 & 1724.76 & 1775.34 \\ \hline
% Baseline \\ (w/o WD) & 1582.40 & 2249.88 & 1724.76 & 1775.34 \\ \hline
CSI & \textbf{1743.63} & \textbf{2377.82} & \textbf{1952.63} & \textbf{1796.92} \\ \hline
\end{tabular}
\end{center}
\end{table}

\section{CONCLUSION}
In this work, we introduce CSI, a flexible framework that enables legged robots to acquire a wide range of controllable and diverse motion skills directly from demonstration motion data. This technology enables the rapid integration of multiple motion skills into a single controller. We believe this capability is advantageous for applications that require legged robots to possess a diverse set of skills. One future work is to deploy our work on real legged robots to validate its feasibility in real robotics applications. Another future work will focus on implementing more detailed motion skill control, such as controlling the direction or the velocity of motion skills. This will further improve the usefulness of our approach.

% Another meaningful direction is to introduce multi-task reinforcement learning, where additional control commands, such as the robot's heading or velocity, are added during the training process. 


% References are important to the reader; therefore, each citation must be complete and correct. If at all possible, references should be commonly available publications.
\bibliographystyle{IEEEtran}
% \bibliography{ref}  % .bib


%%%%%%%%%%%%%%%%%%%%%%%%%%%%%%%%%%%%%%%%%%%%%%%%%%%%%%%%%%%%%%%%%%%%%%%%%%%%%%%%

%%%%%%%%%%%%%%%%%%%%%%%%%%%%%%%%%%%%%%%%%%%%%%%%%%%%%%%%%%%%%%%%%%%%%%%%%%%%%%%%

%%%%%%%%%%%%%%%%%%%%%%%%%%%%%%%%%%%%%%%%%%%%%%%%%%%%%%%%%%%%%%%%%%%%%%%%%%%%%%%%
% \clearpage
\section*{APPENDIX}

\begin{table}[h]
\caption{Detailed statistics of motion capture dataset used in different robots and tasks.}
\label{table:dataset}
\begin{center}
\begin{tabular}{|c|c|c|c|}
\hline
\textbf{Robot}              & \textbf{Task}                  & \textbf{Motion}     & \textbf{Duration} \\ \hline
\multirow{17}{*}{BRUCE}     & \multirow{10}{*}{H-Locomotion} & Walk Forward        & 2.55s             \\ \cline{3-4} 
                            &                                & Walk Backward       & 2.32s             \\ \cline{3-4} 
                            &                                & Walk Rightward      & 2.32s             \\ \cline{3-4} 
                            &                                & Walk Leftward       & 2.98s             \\ \cline{3-4} 
                            &                                & Turn Right          & 1.48s             \\ \cline{3-4} 
                            &                                & Turn Left           & 1.15s             \\ \cline{3-4} 
                            &                                & Turn Right in Place & 1.15s             \\ \cline{3-4} 
                            &                                & Turn Left in Place  & 1.65s             \\ \cline{3-4} 
                            &                                & Run                 & 0.83s             \\ \cline{3-4} 
                            &                                & Jump                & 1.65s             \\ \cline{2-4} 
                            & \multirow{7}{*}{H-WalkStyle}   & Walk Briskly        & 2.10s             \\ \cline{3-4} 
                            &                                & Walk Hobble         & 3.65s             \\ \cline{3-4} 
                            &                                & Walk Stealthily     & 4.15s             \\ \cline{3-4} 
                            &                                & Zombie              & 4.98s             \\ \cline{3-4} 
                            &                                & Walk Happily        & 4.98s             \\ \cline{3-4} 
                            &                                & Marching            & 4.98s             \\ \cline{3-4} 
                            &                                & Walk Slowly         & 3.65s             \\ \hline
\multirow{5}{*}{AlienGo}    & \multirow{6}{*}{Q-Locomotion}  & Pace                & 0.64s             \\ \cline{3-4} 
                            &                                & Trot                & 0.54s             \\ \cline{3-4} 
                            &                                & Pace Backward       & 0.64s             \\ \cline{3-4} 
                            &                                & Trot Backward       & 0.54s             \\ \cline{3-4}
                            &                                & Turn Left           & 0.64s             \\ \cline{3-4} 
                            &                                & Turn Right          & 0.76s             \\ \hline
\multirow{10}{*}{H1}        & \multirow{8}{*}{H-Interact}    & Dance               & 5.02s            \\ \cline{3-4} 
                            &                                & Hand Shake          & 4.82s             \\ \cline{3-4} 
                            &                                & Stand Still         & 4.72s             \\ \cline{3-4} 
                            &                                & Walk Forward        & 2.55s             \\ \cline{3-4} 
                            &                                & Wave Hello          & 5.38s             \\ \cline{3-4}
                            &                                & Walk Proudly        & 5.22s            \\ \cline{3-4} 
                            &                                & Walk Happily        & 3.12s            \\ \cline{3-4}
                            &                                & Zombie              & 4.93s            \\ \hline
\end{tabular}
\end{center}
\end{table}

\subsection{Hyperparameters} \label{training settings}
Adam is used as optimizer for the policy, the value function, and the discriminator in this work, with a fixed learning rate during training. Detailed hyperparameter settings for CSI are shown in \makebox{TABLE~\ref{table:hyperparameter}}. We find that these hyperparameter combinations are suitable for training all tasks in our experiment.

\begin{table}[htb]
\caption{Hyperparameters of CSI}
\begin{center}
\begin{tabular}{|c|c|}
\hline
\label{table:hyperparameter}
\textbf{Parameters}               & \textbf{Value}              \\ \hline
Style-Reward Weight               & 1.0                         \\ \hline
Conditional Imitation Loss Weight & 1.0                         \\ \hline
Condition Aware Loss Weight       & 1.0                         \\ \hline
Weight Decay Loss Weight          & \multicolumn{1}{l|}{0.0001} \\ \hline
Gradient Penalty Weight           & 5.0                         \\ \hline
DoF Velocity Penalty Weight       & -1e-4                       \\ \hline
Action Rate Penalty Weight        & -1e-2                       \\ \hline
Energy Penalty Weight             & -2e-5                       \\ \hline
Torque Penalty Weight             & -1e-4                       \\ \hline
Adjust Ratio                      & 0.5                         \\ \hline
Discriminator Batch Size          & 512                         \\ \hline
MiniBatch Size                    & 32768                       \\ \hline
Learning Rate(for all network)    & 5e-5                        \\ \hline
Discount Factor                   & 0.95                        \\ \hline
Discriminator Replay Buffer Size  & 1e6                         \\ \hline
PPO Clip Threshold                & 0.2                         \\ \hline
GAE                               & 0.95                        \\ \hline
\end{tabular}
\end{center}
\end{table}

\subsection{Baseline Settings} \label{settings}
For a fair comparison, all baselines share the same observation space and action space, and for the same robot, the PD controller parameters used are also identical. In addition, the policies, the value functions, and discriminators used in all baselines are set to the same parameter sizes. Some specific training settings for CAMP and CALM are briefly described below.

\textbf{CAMP} The overall architecture of CAMP is similar to that of AMP~\cite{2021-TOG-AMP}, but one-hot coding of motion skill labels is added to the observation inputs, and the training objective of the discriminator is modified to a conditional loss:
\begin{equation}
\begin{split}
    L = &-E_{(s_{t}, s_{t+1}) \in d^{M}}[D(s_{t}, s_{t+1} | c) - 1]^2 \\
    &- E_{(s_{t}, s_{t+1}) \in d^{\pi}} [D(s_{t}, s_{t+1} | c) + 1]^2 \\
    &+ \omega_{gp} E_{(s_{t}, s_{t+1}) \in d^{M}}[\left\| \triangledown D(s_{t}, s_{t+1}) \right\|^{2}_{2}]
\end{split}
\end{equation}
Accordingly, conditional style reward in CAMP is defined as:
\begin{equation}
    r_{s} = max[0, 1 - 0.25 (D(s_{t}, s_{t+1} | c) - 1)^2]
\end{equation}
Finally, the hyperparameter settings used for CAMP are the same as that of CSI.

\begin{table}[htb]
\caption{Hyperparameters of CALM}
\begin{center}
\begin{tabular}{|c|c|}
\hline
\label{table:calm_hyperparameter}
\textbf{Parameters} & \textbf{Value} \\ \hline
Motion Encoder MLP Size & {[}1024, 512{]} \\ \hline
Policy and Value Function MLP Size & {[}256, 256{]} \\ \hline
Conditional Discriminator MLP Size & {[}256, 256{]} \\ \hline
Encoder Obs Steps & 60 \\ \hline
Learning Rate & 2e-5 \\ \hline
Condition Style Reward Weight & 1.0 \\ \hline
Encoder Regularization Coeff & 0.1 \\ \hline
Latent Dimension & 64 \\ \hline
Discriminator Batch Size & 512 \\ \hline
MiniBatch Size & 16384 \\ \hline
Gradient Penalty Weight & 5.0 \\ \hline
Discount Factor & 0.99 \\ \hline
GAE & 0.95 \\ \hline
PPO Clip Threshold & 0.2 \\ \hline
\end{tabular}
\end{center}
\end{table}

\textbf{CALM} With motion encoder and latent space, after the pre-training stage, the controller trained by CALM~\cite{tessler2023calm} can control a specific integrated skill by providing a short clip of corresponding reference motion data as motion encoder's input. Therefore, in our experimental setup, baseline CALM is only subjected to the pre-training stage. Except for some hyperparameter modifications, we have preserved as much as possible the default settings in the open-source code of CALM. Some key hyperparameter settings used for CALM are shown in \makebox{TABLE~\ref{table:calm_hyperparameter}}. It is also worth noting that the motion state representation used for the CALM's motion encoder is the same as that used for its discriminator. Finally, for a fair comparison, we adjust all the observations of CALM's policy, value function, and discriminator to be the same as other baselines (except condition input).

\subsection{Scalability Analysis}
\begin{figure}[htb]
    \centering
    \includegraphics[width=0.9\linewidth]{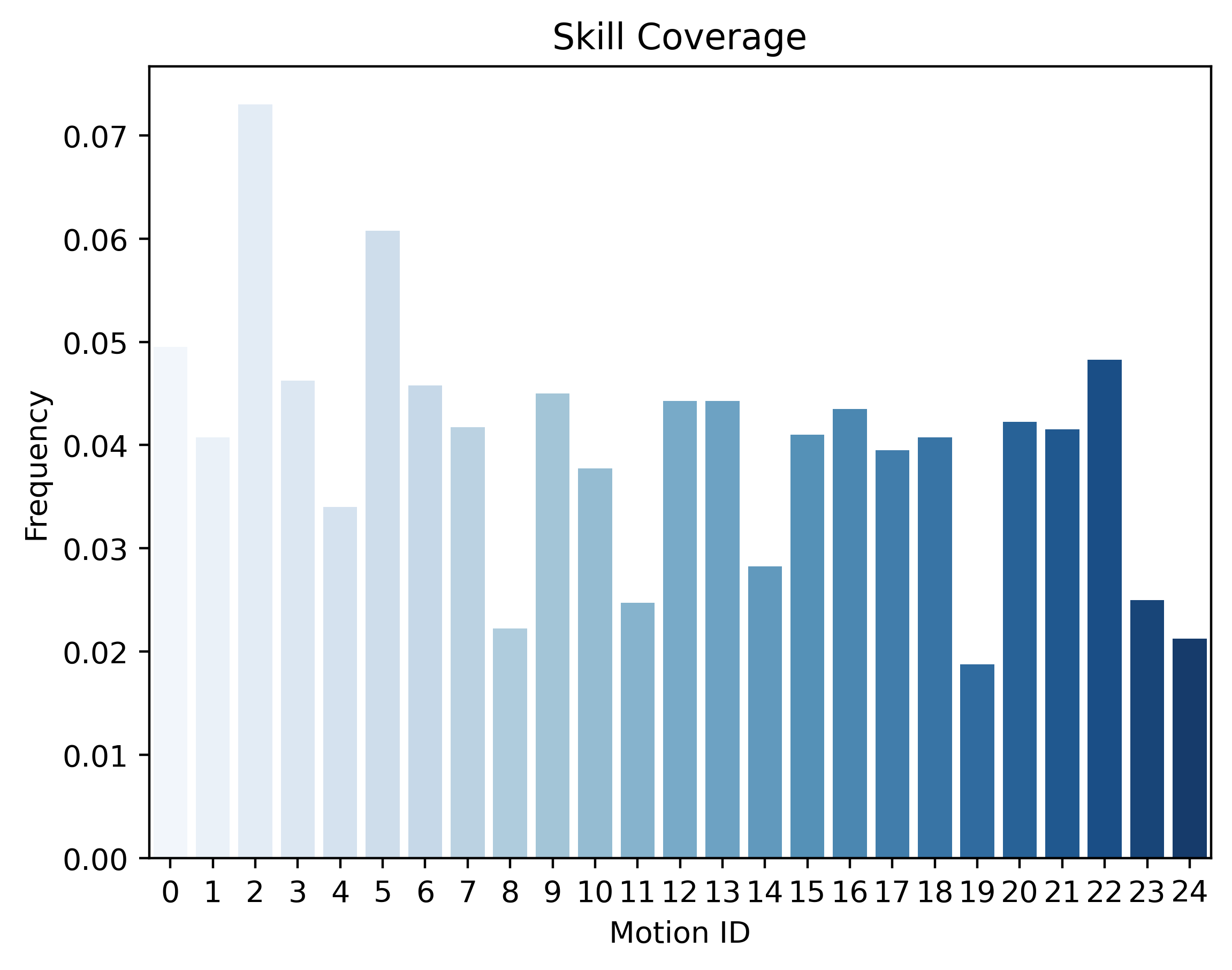}
    \caption{CSI demonstrates great scalability even if the number of skills to be integrated increases significantly.}
    \label{fig:scalability}
\end{figure}

This section demonstrates the scalability of CSI when the number of reference motion skills increases. We significantly increase the number of motion skills required to be integrated into H1 by constructing a dataset containing $25$ different reference motion skills. This dataset contains all the reference datasets used by the humanoid robots in this paper, as well as a set of similar mirror motion skills (\textit{wave left hand} and \textit{wave right hand}, \textit{left leg kicking} and \textit{right leg kicking}, \textit{left hand shake} and \textit{right hand shake}). Additionally, we keep the size of each network unchanged and only extend the training duration to $2.5$ times of the original. \makebox{Fig.~\ref{fig:scalability}} illustrates the coverage of motion skills. CSI demonstrates excellent scalability by mastering all motion skills despite a significant increase in the number of skills to be integrated and the inclusion of a set of similar motion skills.

\subsection{Generality  Analysis}
\begin{figure*}[htb]
\subfigure[AlienGo: \textbf{(1)Trot, (2)Pace, (3)Turn}]{
    \begin{minipage}[t]{0.97\linewidth}
        \centering
        \includegraphics[width=1\linewidth]{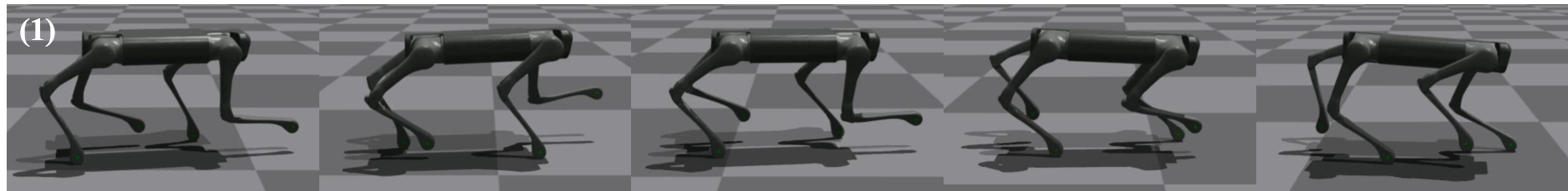}\vspace{0.5mm}
        \includegraphics[width=1\linewidth]{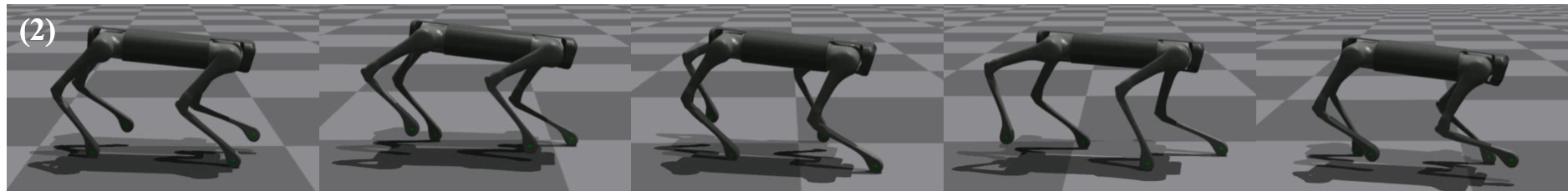}\vspace{0.5mm}
        \includegraphics[width=1\linewidth]{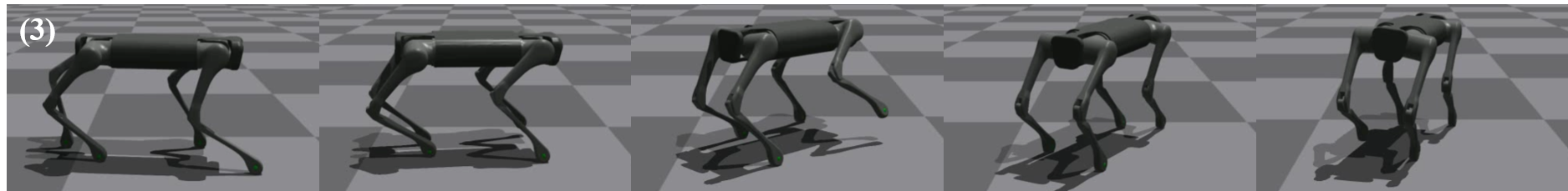}\vspace{1mm}
    \end{minipage}
}

\vspace{-2mm}

\subfigure[BRUCE: \textbf{(1)Zombie, (2)Marching, (3)Walk Happily, (4)Walk Stealthily}]{
    \begin{minipage}[t]{0.49\linewidth}
        \centering
        \includegraphics[width=1\linewidth]{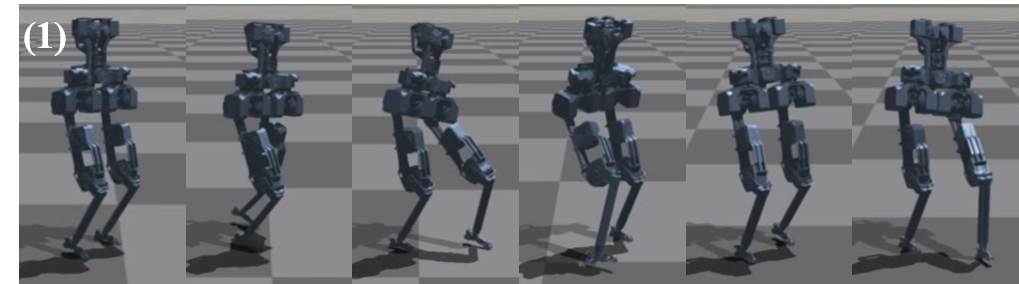}\vspace{0.5mm}
        \includegraphics[width=1\linewidth]{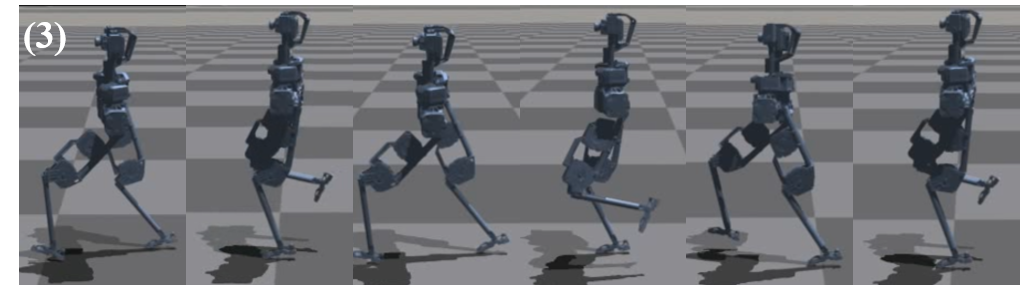}\vspace{1mm}
    \end{minipage}
    \hspace{-1.5mm}
    \begin{minipage}[t]{0.49\linewidth}
        \centering
        \includegraphics[width=1\linewidth]{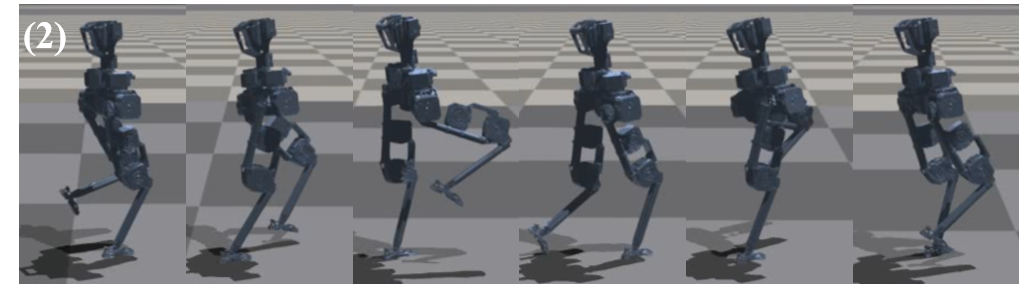}\vspace{0.5mm}
        \includegraphics[width=1\linewidth]{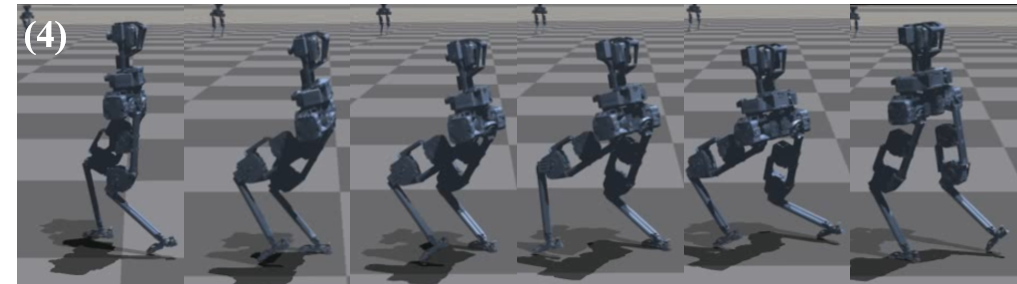}\vspace{1mm}
    \end{minipage}
}

\vspace{-2mm}

\subfigure[H1: \textbf{(1)Wave Hello, (2)Walk Proudly, (3)Dance, (4)Walk Forward}]{
    \begin{minipage}[t]{0.49\linewidth}
        \centering
        \includegraphics[width=1\linewidth]{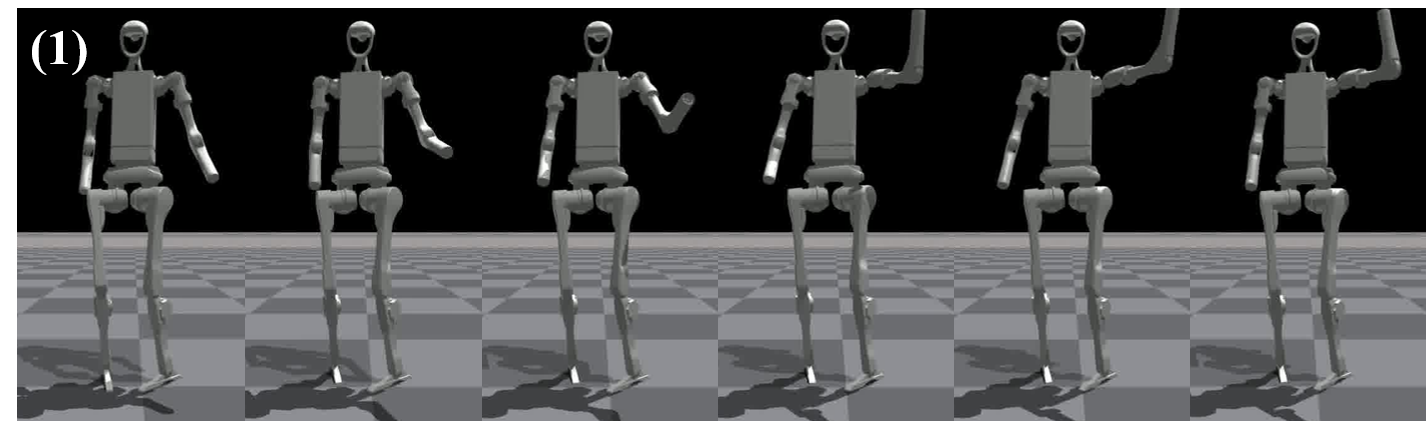}\vspace{0.5mm}
        \includegraphics[width=1\linewidth]{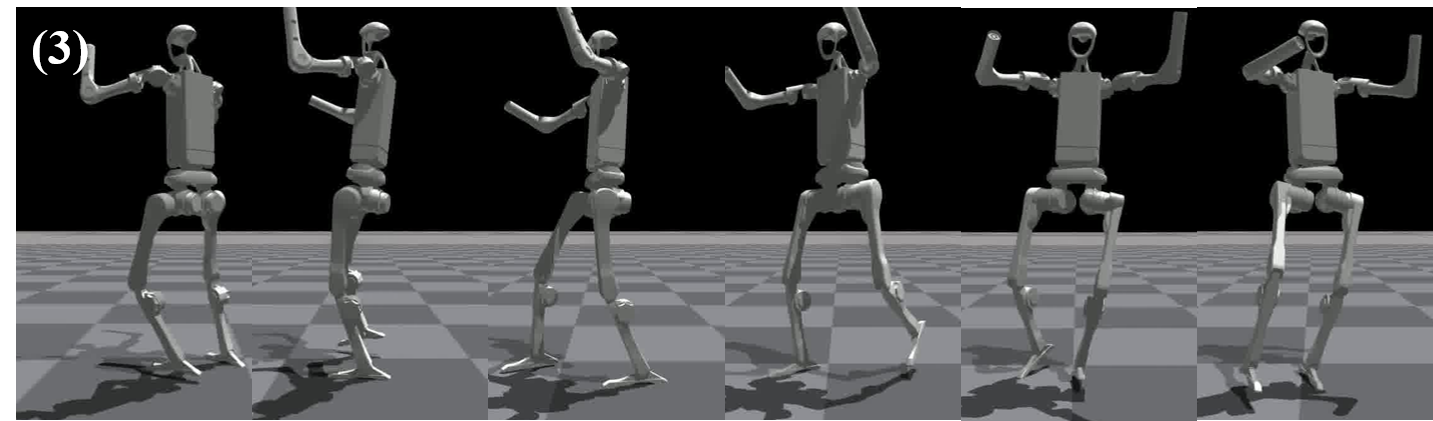}\vspace{1mm}
    \end{minipage}
    \hspace{-2mm}
    \begin{minipage}[t]{0.49\linewidth}
        \centering
        \includegraphics[width=1\linewidth]{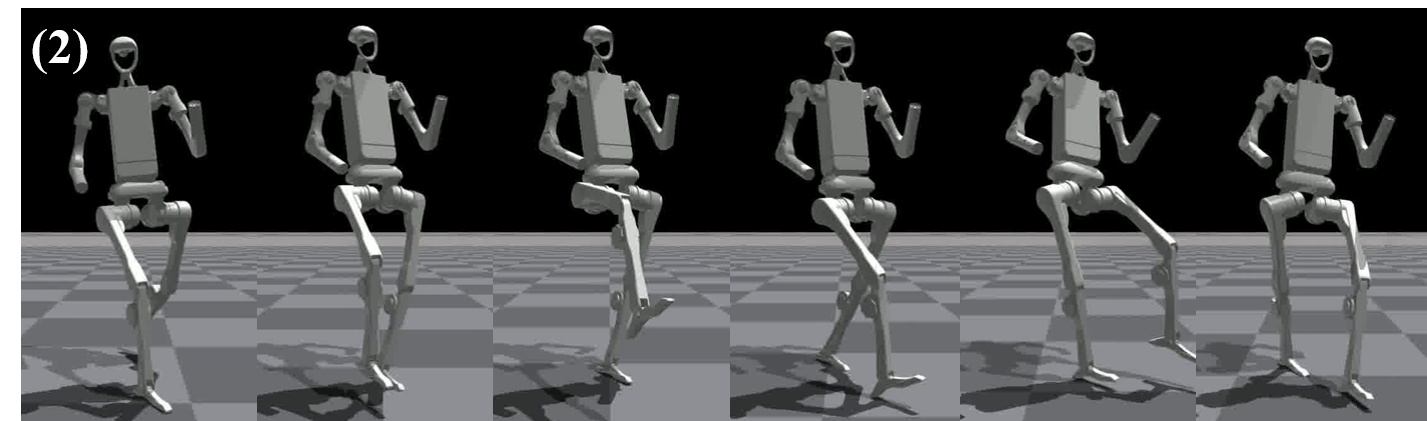}\vspace{0.5mm}
        \includegraphics[width=1\linewidth]{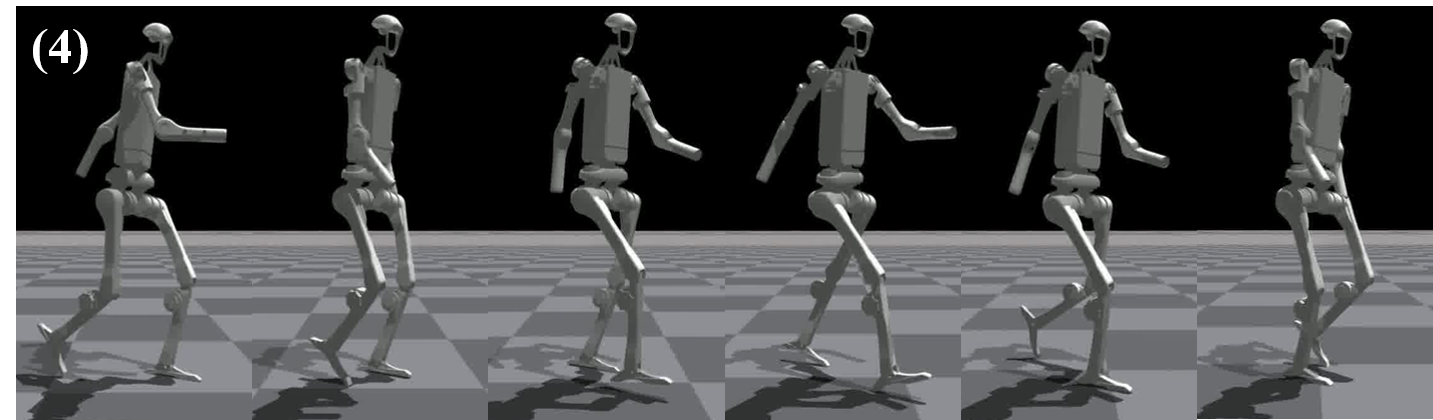}\vspace{1mm}
    \end{minipage}
}

\vspace{-1mm}

\caption{CSI can be applied to different legged robots for versatile and flexible multi-skill integration.}
\label{fig:generality}
\end{figure*}

This section briefly analyzes the generality of CSI. As shown in \makebox{Fig.~\ref{fig:generality}}, CSI can be applied to different humanoid robots BRUCE and H1 as well as quadruped robot AlienGo, and they share the same training process. For other humanoid robots, CSI is also theoretically applicable, but the difficulty lies in how to kinematically retarget the motion capture data for the robot to obtain the corresponding reference motion dataset. Additionally, the motion capture data of quadrupedal creatures is more difficult to obtain than that of human beings, so the source of the reference motion skill is a major obstacle to the application of CSI on quadruped robots. In this regard, we believe that future work is necessary to further expand more sources of reference motions, such as extracting reference motions from videos~\cite{yuan2022glamr, cai2024smpler}, or obtaining reference motions from generative models~\cite{jiang2024motiongpt, tevet2023human}. These expanded reference motion data sources will further enhance the generality of CSI.

Another point is that CSI requires less correlation between motion skills that need to be integrated. The reference motion skills that can be integrated by previous work~\cite{escontrela2022adversarial, li2023learningterrain, tang2023humanmimic} are limited by the task objective. For example, in locomotion tasks, policies can usually only integrate task-relevant motion skills such as walking, standing, steering, and running. Integrating additional jumping skills would require further adjustments to the task objective. CSI does not have any obvious relevance requirements for this, as shown in \makebox{Fig.~\ref{fig:generality}}, where similar motion skills, such as \textit{trot} and \textit{pace}, as well as very different motion skills, such as \textit{dance} and \textit{zombie}, can be integrated, highlighting the generality of CSI for multi-skill integration.

% \subsection{Domain randomization} 
% Overcoming the gap between simulation and reality has long been a difficulty standing in the way of deploying a neural network based controller to a real robot. To facilitate the future transfer of trained policy to the real robot, we apply domain randomization~\cite{tobin2017domain}. Specifically, during the training process, we randomize physical factors such as the robot's mass and ground friction coefficient, \makebox{TABLE~\ref{table:domain}} demonstrates the detailed randomization objects and the corresponding ranges. The actual randomized values will be obtained by uniform sampling within the range. Additionally, to improve the robot's robustness for external interference, we add an extra sampled force vector to the randomly selected robot's body parts during training to disturb the robot every $5$ seconds. 

\subsection{Initialization}
We adopt a mixed initialization strategy to accelerate motion skill learning. At the beginning of each episode, a set of initialization state and skill label pairs ${s, c}$ are selected to initialize each agent, where $70\%$ of $s$ are sampled from $d^{M}$, the remaining $30\%$ is set to default state, and all skill labels $c$ are randomly sampled from skill label space $C$. Note that $c$ and $s$ may derive from different reference motion skills, which would be beneficial for learning to switch between different motion skills.

\subsection{Skill labelling}
In this work, we crop all the reference motion clips used so that they contain only one motion skill. We then artificially assign a unique skill label to each clip. Actually, for unstructured reference motion data, such as motion capture data clips with a mixture of multiple motion skills, skill labels can be considered to be acquired by pre-trained skeleton-based action recognition networks like~\cite{duan2022revisiting, duan2022pyskl}, and similar practice has been reported in~\cite{dou2023case}.

% \begin{table}[t]
% % \setlength{\abovecaptionskip}{-0.1cm}
% \caption{Randomization Simulation Parameters}
% \label{table:domain}
% \begin{center}
% \begin{tabular}{ccc}
% \toprule
% Parameters & Lower Bound & Upper Bound\\
% \midrule
% terrain friction & 0.2 & 1.0 \\
% added mass & -1.2 kg & 1.2 kg \\
% added inertia & 0.5 & 1.5 \\
% joint added damping scale & 0. & 0.05 \\
% joint added friction scale & 0. & 0.05 \\
% PD gain scale range & 0.9 & 1.1 \\
% \bottomrule
% \end{tabular}
% \end{center}
% % \vspace{-0.5cm}
% \end{table}

% \section*{ACKNOWLEDGMENT}

% The preferred spelling of the word ÒacknowledgmentÓ in America is without an ÒeÓ after the ÒgÓ. Avoid the stilted expression, ÒOne of us (R. B. G.) thanks . . .Ó  Instead, try ÒR. B. G. thanksÓ. Put sponsor acknowledgments in the unnumbered footnote on the first page.

%%%%%%%%%%%%%%%%%%%%%%%%%%%%%%%%%%%%%%%%%%%%%%%%%%%%%%%%%%%%%%%%%%%%%%%%%%%%%%%%

\end{document}